%% file: main.tex
\documentclass[conference]{IEEEtran}
\IEEEoverridecommandlockouts
\usepackage{cite}
\usepackage{amsmath,amssymb,amsfonts}
\usepackage{algorithmic}
\usepackage{graphicx}
\usepackage{textcomp}
\usepackage{xcolor}
\usepackage{booktabs}
\usepackage{hyperref}
\usepackage{xurl}
\usepackage{multirow}

\usepackage{multicol}
\usepackage{verbatim}
\usepackage{array}

\def\BibTeX{{\rm B\kern-.05em{\sc i\kern-.025em b}\kern-.08em
    T\kern-.1667em\lower.7ex\hbox{E}\kern-.125emX}}
\begin{document}

\title{Progressive Feature Upgrade in Semi-supervised Learning on Tabular Domain*\\
}
 \author{\IEEEauthorblockN{1\textsuperscript{st} Morteza Mohammady Gharasuie}
    \IEEEauthorblockA{\textit{Department of Computer Science} \\
    \textit{Old Dominion University}\\
    Norfolk, USA \\
    mmoha014@odu.edu}
    
 \and
    \IEEEauthorblockN{2\textsuperscript{nd} Fenjiao Wang}
    \IEEEauthorblockA{\textit{Department of Computer Science} \\
    \textit{Old Dominion University}\\
    Norfolk, USA \\
    f1wang@odu.edu}}


\maketitle

\input{abstr}


\input{intro}
\input{relate.tex}
\input{methodology}

\input{experiment.tex}
\input{conclusion}

\bibliographystyle{IEEEtran}
\bibliography{references}

\appendix
\input{supplementary/supp}

\end{document}

%% file: abstr.tex
  \begin{abstract}
 Recent semi-supervised and self-supervised methods have shown great success in the image and text domain by utilizing augmentation techniques. 
 Despite such success, it is not easy to transfer this success to tabular domains. 
 It is not easy to adapt domain-specific transformations from image and language to tabular data due to mixing of different data types (continuous data and categorical data) in the tabular domain. There are a few semi-supervised works on the tabular domain that have focused on proposing new augmentation techniques for tabular data. These approaches may have shown some improvement on datasets with low-cardinality in categorical data. However, the fundamental challenges have not been tackled. The proposed methods either do not apply to datasets with high-cardinality or do not use an efficient encoding of categorical data. We propose using conditional probability representation and an efficient progressively feature upgrading framework to effectively learn representations for tabular data in semi-supervised applications. The extensive experiments show superior performance of the proposed framework and the potential application in semi-supervised settings.
 
  \end{abstract}

%% file: intro.tex
\section{Introduction}
Since the major breakthrough in the ImageNet Large Scale Visual Recognition Challenge, deep learning has attracted much attention due to its superior performance in many applications, e.g. Speech Recognition, Computer Vision, and Natural Language Processing. 
Such great progress is largely driven by enormous datasets. 
Collecting and labeling such an enormous dataset is expensive, time-consuming, and often impossible.

Recently, semi-supervised learning \cite{laine2016temporal, liu2021semi,sohn2020fixmatch} has gained a lot of attention due to superior performance in the image and language domain.  
Semi-supervised learning aims to leverage a small amount of labeled data as well as a huge amount of unlabeled data to perform learning tasks (classification and regression).
Recently, researchers propose different augmentation techniques and regularizations (consistency regularization) in semi-supervised learning approaches \cite{simCLR}. 
Often, these augmentation techniques are domain specific.
Take image domain augmentation as an example, these augmentation techniques 
help create images that can explicitly cover various perturbations/variances (viewpoint, lighting, occlusion, background in image domain) to challenge the learned model to better handle these difficult cases.
Theoretically, the data augmentation technique in semi-supervised learning can help increase the generalization ability of the trained models by reducing the overfitting and expanding the decision boundary of the models. 

However, the success of semi-supervised learning approach on image and language domain can hardly be transferred to tabular domain. Because, the success of augmentation-based semi-supervised learning algorithms heavily relies on the spatial or semantic structure of image or language data. 
For other data like tabular data which does not exhibit any explicit structure, the semi-supervised learning becomes much more challenging. 
The suspected reason is that augmentation technique like Mixup (\cite{zhang2017}) is usually a convex combination of the original samples. These augmentation techniques may only work well if the original data manifold is likely convex too. Meanwhile, tabular data is likely not convex.
Directly applying augmentation techniques that have been used in image and language domain to tabular domain can easily create out-of-distribution samples which may even hurt the learning process \cite{yoon2020vime}. 


One straightforward idea towards solving semi-supervised learning in the tabular domain is to develop some customized augmentation and loss function for tabular data.
This route is challenging because of the following reasons.
Tabular data does not exhibit any explicit structure, coming up with any suitable augmentation is not an easy task. 
Also, tabular data usually contains both categorical data which is discrete and numerical data which is continuous. 
Mixing different data types altogether creates a severe challenge for the recent semi-supervised learning approaches.
Some recent works on semi-supervised learning to tabular data have focused on proposing new augmentation operations suitable for tabular data or working on the latent space.
However, even the new augmentation technique can not fundamentally solve the problem due to the high-cardinality of categorical data and inefficient representation.

We propose to change the representation of the tabular data especially categorical data from common approach (one-hot encoding, label encoding) to less known approach (conditional probability representation) to enjoy several unique benefits mentioned in the next paragraph. 
We argue that one should take a step back and carefully examine how tabular data and especially how categorical data can be represented in the semi-supervised learning problem.
Representation of the data is one important aspect that can easily be ignored.
Our experiment in Table~\ref{tab:compare_encodings} also shows that the choice of representation for categorical data may greatly impact the performance.
Also, such impact may even be agnostic to what follow-up model is being used.


We propose utilizing conditional probability representation (CPR) for semi-supervised learning in the tabular domain. 
CPR maps individual values of categorical feature to the probability estimate or the expected value of the target attribute.
In another word, it computes the likelihood of a specific categorical value leading to a specific label.
It has many unique benefits compared to other representations (one-hot encoding, label encoding). 
Firstly, it is an efficient representation in terms of how many bits are used to represent the feature, especially for high-cardinality categorical data. 
The reason being that the number of dimensions of CPR does not depend on the cardinality of the categorical feature. It only depends on the number of target labels. 
Secondly, label information has been baked into the representation. 
Label information is critical for semi-supervised learning algorithm. If one can inject label information into the feature, it may be easier for the model to learn meaningful representations.
More importantly, it opens the door for utilizing pseudo-labels (predicted labels) in a novel way (constructing the feature).  
Thirdly, compared to other representations, CPR is closer to the numerical features, since it uses conditional probabilities as features. This property may open the door for better enabling leveraging various existing augmentation techniques for tabular data.

Beside employing CPR as feature representation for categorical data, more importantly, we propose to progressively upgrade the CPR during model training by leveraging the pseudo-labels.
Instead of only using true labels to construct CPR, we propose to use pseudo-labels to update the CPR during the model training process.
Our initial study shows that even if we don't change how much data is used for training, but only increase the amount of data used for constructing the CPR, the prediction accuracy can be hugely boosted (Table~\ref{tab:updatePolicy}).
Pseudo-labels are defined as predicted class labels for unlabeled data. 
Self-training algorithm utilizes pseudo-labels of unlabeled data to continue improving the model. 
It is just one way of utilizing pseudo-labels which is treating pseudo-labels as if they are the ground-truth labels in training the model. 
However, pseudo-labels can also be used in another way which is to update the CPR and then to influence the model training process. 
With model being trained progressively, more accurate pseudo-labels will help generate more accurate representations for learning in the ``feedback loop''.

We propose a framework that can progressively upgrade the CPR representation.
The proposed framework is flexible in the sense that it can act as an add-on component to the existing semi-supervised learning frameworks. 
For this framework to work, there needs to be a component that can produce pseudo-labels. This condition is not hard to satisfy.
Commonly, the semi-supervised learning algorithms always contain a component (predictor or label propagation) that can produce pseudo-labels. 
The pseudo-labels provided by these components can then be used to upgrade the CPR. 
One clear benefit is that even if certain category values do not exist in the labeled dataset, representation for those categories can still be calculated because of employing the generated pseudo-labels. 
Pseudo-labels being used for updating the CPR representation are not 100\% correct, it may introduce additional noise to the model training process. We propose several refinement mechanisms to alleviate such issue by selecting only the pseudo-labels with high confidence that they are the correct label.

The main contribution of our paper can be summarized as follows.
\begin{itemize}
\item We propose using conditional probability representation for high-cardinality categorical data for efficient representation. To the best of our knowledge, our work is the first work that uses an encoding different from one-hot encoding for tabular domain in semi-supervised learning. The proposed framework can also be extended to other encoding method such as target encoding (\cite{TE2001}) which also bakes the target label into the representation. 
\item We propose novel feature upgrading framework by leveraging pseudo-labels. To the best of our knowledge, we are the first paper to propose using pesudo-labels to update the CPR for categorical data in semi-supervised learning. 
\item The proposed framework is flexible and complementary which can be easily embedded into the existing semi-supervised learning algorithms to boost the learning performance.
\item We demonstrate the superior performance of the proposed framework in extensive experiments.  
The robustness of the proposed framework has been testified by superior performance on two different semi-supervised algorithms and three different tabular datasets. 
\end{itemize}

The paper is organized as follows. In Section 2, we provide a review of the related semi-supervised learning algorithms and works on the representation for categorical data, noting the importance of using and updating the conditional probability representation. In Section 3, we describe the details of the proposed framework. Then, we present the quantitative and qualitative experimental results in Section 4. Finally, Section 5 gives the conclusion.

%% file: relate.tex
\section{Related Works}
\subsection{Semi-supervised Learning}
Semi-supervised learning in general is attempting to improve the performance of the learning algorithms by utilizing both the labeled and unlabeled data, such that the resulted classifier is better than the trained classifier on just labeled data ~\cite{van2020survey}. Semi-supervised learning has shown considerable progress in the language and image domains in recent years. Most of these works resulted from the consistency regularization and pseudo-labeling on the unlabeled data.

\textbf{Consistency Regularization:} The consistency regularization uses the different perturbation of an input sample and tries to enforce the same prediction for all the perturbations. These perturbations can be applied on either different epochs ~\cite{laine2016temporal, liu2021semi} or same epoch ~\cite{sohn2020fixmatch,sajjadi2016regularization,luo2018smooth}. Also, the perturbation can be applied in the network (dropout, random max-pooling),  the input space \cite{simCLR,mixmatch2019,verma2019interpolation}, and the latent space ~\cite{sajjadi2016regularization,cheung2020modals,kumar2019closer}. 

\textbf{Pseudo-labeling and self-training:} The goal of pseudo-labeling \cite{lee2013pseudo,shi2018transductive} and self-training\cite{mcclosky2006effective} refers to a classical semi-supervised approach where the model is being trained on the labeled and unlabeled samples using labels and pseudo-labels associated with the unlabeled samples. The self-training \cite{xie2020self,haase2021iterative} has recently shown improved performance over supervised counterpart. 
Some works \cite{mukherjee2020uncertainty,rizve2021defense} use calibration and uncertainty of predictions for the selection of samples to improve the pseudo-label selections. Also, disagreement-based models \cite{zhou2010semi} use multiple learning algorithms and exploit the disagreement during the learning process to filter out wrong pseudo-labels.

\subsection{Representation for tabular data}
It is hard to transfer the semi-supervised learning algorithms proposed in image and language domains to tabular domain. Unlike image and language domains, tabular domain is a combination of different data types (numeric and non-numeric data). 
The non-numeric data can be unordered categories with a fixed set of possible values. General idea for processing tabular domain data is to encode it to the numerical representation to better consumed by machine learning algorithms. The classic approach to encode categorical variables is one-hot encoding, which is not suitable for high-cardinality categories due to generating high-dimensional vectors. This is a big problem in large datasets, which might have a very large number of categories, posing computational problems \cite{cerda2020encoding}. Despite the existence of data cleaning 
~\cite{pyle1999data,rahm2000data} and similarity encoding techniques ~\cite{cerda2018similarity}, it is hard to tackle the problem of high cardinality. For this purpose,  Cerdar P., \& Varoquaux G. ~\cite{cerda2020encoding} proposed a scalable encoding method for string categories using min-hash encoding and Gamma-Poisson factorization. They also proposed a similarity encoding technique ~\cite{cerda2018similarity} to encode dirty, non-curated categorical data. Also, Slakey A. et al. \cite{slakey2019encoding} proposed a CBM encoding approach to represent categorical features in low dimension. 

\subsection{Semi-supervised Learning in tabular domain}
Recent advances in semi-supervised learning using deep networks have been applied to the tabular domain in some works. Darabi S. et al. ~\cite{darabi2021contrastive} proposed a semi-supervised framework for the tabular domain, called “contrastiveMixup”. They applied the one-hot encoding on raw categorical data, and the Mixup operation in the latent space among samples with the same labels and pseudo-labels on labeled and unlabeled data respectively.  Supervised contrastive learning and mixup augmentation in the latent space are used to push the samples with the same label closer to each other in the latent space.  Also Yoon J. et al. ~\cite{yoon2020vime} proposed a semi-supervised method for the tabular domain using consistency loss among perturbed versions of the one-hot encoded input samples. The proposed framework introduced an augmentation technique in the input space, which is used to learn the latent space representation using an autoencoder. Then, the pre-trained encoder of autoencoder is used in the semi-supervised setting to learn from labeled and unlabeled samples using consistency losses and perturbation of samples. Also, Ucar T. Et al. ~\cite{subtab2021} introduced a new framework that turns the tabular data into a multi-view representation learning. They claim that reconstructing the data from subsets of features captures a better latent representation rather than reconstructing the corrupted version of input in an autoencoder. Beside of using subset of features, they utilize the contrastive loss, the distance loss among the latent space of subsets, and the different augmentations in the input space to get the best performance.

Recent works only use one-hot encoding and test on small datasets containing low-cardinality categorical features. 
The existing works using one-hot encoding on these datasets usually work fine because the one-hot encoded features are not big. But in case of big datasets with high cardinality, the one-hot encoding is not the best choice. There are many different encodings and several Python libraries cover them
\footnote{\url{https://contrib.scikit-learn.org/category_encoders/}} and \footnote{\url{https://dirty-cat.github.io}}. The different encodings have different performance on different datasets and it motivate us to think about using other encoding methods in semi-supervised learning. We consider using a conditional probability representation (CPR) that uses label of data for creating  a numerical and continuous representation of the categorical data. We also propose a new framework that uses the CPR and progressively updates the representation for categorical features.
Intuitively, this representation is more friendly to existing augmentation techniques than other representations (one-hot encoding or label encoding).

To the best of our knowledge, our work is the first work that uses an encoding different from one-hot encoding for tabular domain in semi-supervised learning. In this regard, we considered big datasets in experimental sections to show the effectiveness of our work.

%% file: methodology.tex
\section{Methodology}

In this section, we describes our proposed framework. We  describe the conditional probability representation (CPR), \emph{Update Policy} and \emph{Refinement} methods for the proposed framework. Then, we introduce \emph{progressive VIME} and \emph{Progressive Contrastive Mixup}.

\subsection{Conditional Probability Representation}
In case of big datasets contatining high-cardinality categorical data, the one-hot encoding is not the best choice because of the space consumption and curse of dimensionality problems \cite{verleysen2005curse}. In contrast,  the CPR of the categorical data has a fixed representation w.r.t the number of targets in the classification problem, which creates a compact representation because usually the number of labels is much smaller than the cardinality of the features. Table \ref{tab:compare_encodings} compares the performance of a Multi-layer perceptron (MLP) network on the CPR and two other encodings using Traffic Violations dataset. This experiment shows that the MLP using CPR can outperform the same model while using one-hot encoding or label encoding. 
Please note that different encoding methods may perform differently across different datasets/applications.

\begin{table}
  \caption{Test accuracy on traffic violations dataset based on an MLP with three layers (256D-128D-4D) shows the conditional probability representation leads to better accuracy.}
  \label{tab:compare_encodings}
  \centering
 
  \begin{tabular}{cc}
    \toprule
 
    \textbf{Representation} & \textbf{Test Accuracy}\\
    \midrule
    Label Encoding & 55.3\%\\
    One-hot encoding & 69.23\%\\
    Conditional probability encoding & 73.33\%\\
  \bottomrule
\end{tabular}
\vspace{-3mm}
\end{table}

To define the problem mathematically, let \(X\) be an \(N*M\) matrix with row vectors \(X_n\) and column vectors \(X_m^T\).  Let \(Y\) be  an N-dimensional target vector and \(Y_n\) is the observed value correspond to \(X_n\). Then \(D=(X,Y)\) is a dataset with \(N\) samples where \(Dn=(Xn,Yn)\) is the $n^{\text{th}}$ sample with label target \(Y_n \in \{0,1,...,C\}\). In context of the categorical data problem, column \(X_m^T\) with cardinality \(K_m\) has a domain \(V(m)=\{X_{n,m}\}_n\) containing unique nominal values \(V_m \in 1,...,k_m\). Let \(C\) be the number of the values in the target variable. 
The CPR measures given each  category value in a categorical feature, how likely for this category value lead to different target labels within the dataset.
Therefore, for each categorical feature, a $C$-dimensional vector representation will be produced, where C is the number of target labels in the dataset. Following equation computes the CPR.

$X_{n,m} = [\frac{N_{n,m,1}}{N_{n,m}},\frac{N_{n,m,2}}{N_{n,m}},...,\frac{N_{n,m,C}}{N_{n,m}}] $ 

where $N_{n,m,c}$ is the number of observation of categorical value \(X_{n,m}\) that belongs to the label target \(c \in C\), and  \(N_{n,m}\) is the number of observation of categorical value \(X_{n,m}\) in  \(X_m^T\). The summation of all the conditional probabilities is $1$.

\subsection{Preliminaries}
To present the method, we formulate the semi-supervised problem. Consider a dataset with N samples. There are a small subset of labeled samples \begin{math}D_L=\{(X_n,y_n)\}_{n=1}^{N_L}\end{math}  and a large set of the unlabeled samples \(D_U=\{(x_n)\}_{n=L}^{N_U}\) where \(N = N_U+N_L\). We consider the setting where  \(N_U>>N_L\). The supervised training on labeled samples without learning from unlabeled samples mostly likely causes overfit. The unlabeled samples can be used to improve the generalization of the model to get better accuracy on unseen test samples. For this purpose, we use pseudo-labels and an \emph{Update Policy} for better generalization in the training of the neural network, and also a better representation of data that helps boost the performance.
\subsection{Update Policy}

We design \emph{Update Policy} that incorporates pseudo-labels of the unlabeled samples to update the CPR. 
More samples help generate more efficient representation. Our initial study in Table \ref{tab:updatePolicy} shows that more labeled samples used for generating representation indeed drastically improve the model performance. This study shed a light on our approach that using more ``labeled'' samples may boost the model performance.  
When we use more data with ground truth to generate the representation,
the new representation of data influences the performance and improves
the accuracy. The difference in this study is that we used the
ground truth labels for statistics, however, in reality, pseudo-labels
are being used.

In this approach, we use labeled samples \(D_L\) at first to calculate the conditional probability on categorical features and generate the initial representation of all samples in the dataset (\(D\)). This representation will be updated using both labels for \(D_L\) and pseudo-labels for \(D_U\), and keep on training the semi-supervised model on the updated representation. By updating the representation using more samples (\(D_L\) + \(D_U\)), the model can obtain better generalization since the representation contains more information from the dataset.

\begin{table}
  \caption{The representation using more data influences the performance. Test accuracy on traffic violations dataset based on an  MLP with three layers (256D-128D-4D) shows the effect of using more data. All categorical features are used for the experiment.}
  \label{tab:updatePolicy}
   \centering
  \begin{tabular}{p{0.35\linewidth}p{0.2\linewidth}p{0.25\linewidth}}
    \toprule

    \# \textbf{samples to creating representation} & \# \textbf{samples for training} & \textbf{Test Accuracy}\\

    \midrule
    1024 & 1024 & 72.7\% \\
    102,400 &1024 & 77.32\% \\
  \bottomrule
\end{tabular}
\vspace{-5mm}
\end{table} 

\subsection{Refinement}
\label{subs:refinement}
We propose \emph{refinement} mechanism for handling noise in pseudo-labels. 
It works by filtering out likely incorrect pseudo-labels.
It helps generate more accurate representation in \emph{Update Policy} and improve the performance of the trained model. 
Some methods are introduced to choose more accurate pseudo-labels.
Note that, though how to find more accurate pseudo-labels have been discussed in these papers, our main idea of utilizing CPR and keep updating the representation progressively is different from the methods proposed in these papers.
These works use weight of pseudo-labels in the graph-based label propagation \cite{iscen2019label}, the confidence of pseudo-labels in a classifier \cite{mixmatch2019}, uncertainty weight for each sample \cite{arazo2020pseudo,pl2021defense} or using all pseudo-labels without \emph{refinement} \cite{longtail2021semi}. 

If the label-propagation method is an component in the method, we use measured weights in the label-propagation method for filtering. If the classifier is available, we can use confidence of pseudo-labels in the classifier for filtering. When both label-propagation and classifier are used in the architecture, we propose a mechanism to leverage both components for filtering. Two steps of filtering are used in the proposed mechanism. First, we keep only those pseudo-labels agreed between classifier and label propagation methods. Then, the final pseudo-labels are selected based on a threshold on the measured weights by the label-propagation method. After the second step, the final pseudo-label and its corresponding sample are used by \emph{Update Policy} to update the CPR on \(D\). When one of the components is available in the architecture, the agreement can not be used. Instead, a threshold on the confidence of prediction in the classifier or the weight of pseudo-label in label-propagation is used for filtering.

\begin{figure}[t]
     \centering
      \centering
         \includegraphics[width=0.49\textwidth]{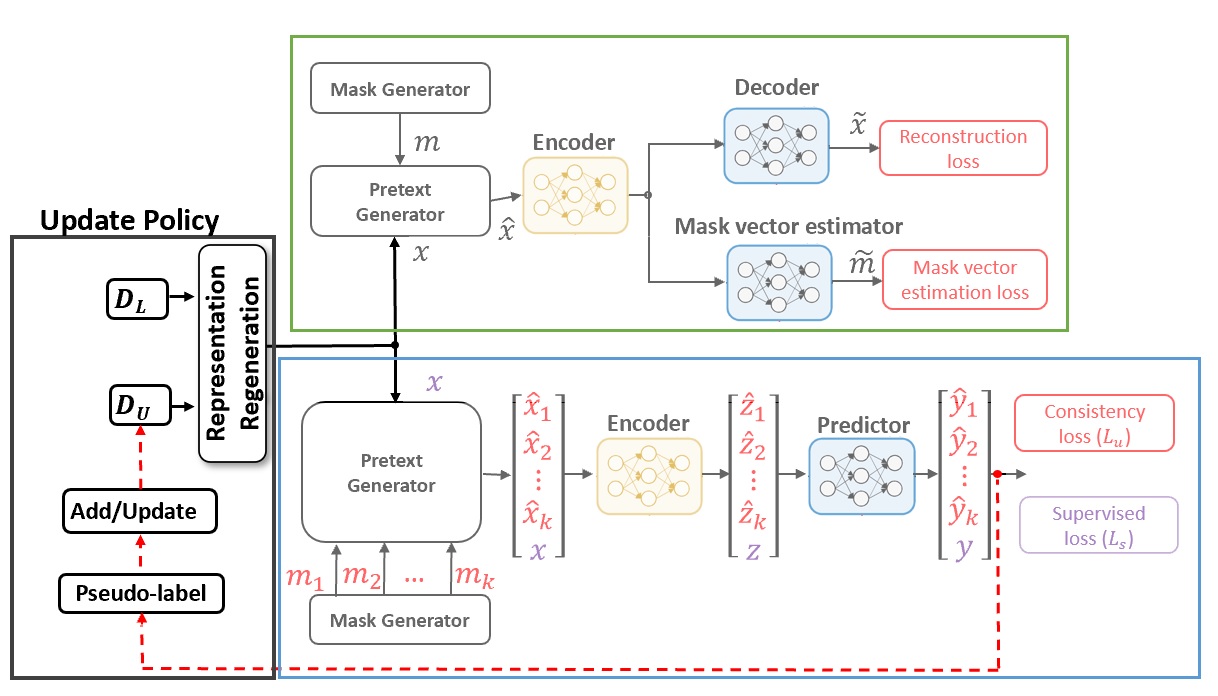}
         \label{fig:curls}
    \vspace*{-5mm}
        \caption{\emph{Progressive VIME} architecture. Green and blue boxes show the first and second steps respectively. The proposed component is illustrated in the black box containing representation regeneration.}
        \vspace*{-3mm}
     \label{fig:VIME}
\end{figure}

\subsection{Progressive Training}
Progressive training refers to updating the input feature representation to train the model. We believe that by updating representation using pseudo-labels, it is possible to train a more effective model. This section shows how \emph{Update Policy} and \emph{Refinement} are used in the progressive architecture. In this regard, the progressive architecture is plugged to two existing semi-supervised learning architectures \cite{darabi2021contrastive,yoon2020vime} that have extended the recent advances in semi-supervised learning to the tabular domain. In the following, we shortly describe the previous works, then expand them to the proposed progressive approach. We utilize  \emph{Update Policy} and \emph{Refinement} for expansion.

\subsubsection{VIME}~\\
Before introducing the proposed \emph{Progressive VIME}, in this section, we first describe \emph{VIME}~\cite{yoon2020vime}. 
\emph{VIME} has two training steps. 
A self-supervised pretext task is proposed in the first step to learn the data representation. Then, consistency regularization is leveraged to fine-tune the prediction.
These steps can be seen in Figure \ref{fig:VIME} inside right upper green box (step 1) and right lower blue box (step 2). 
The black box in Figure \ref{fig:VIME} is not part of the original architecture, it is the component proposed in this paper. It will be described in the next subsection.

In the first step, the representation of data is learned using a denoising autoencoder architecture. The input samples are corrupted by a augmentation method through \emph{mask generator} and \emph{pretext generator} components in both steps before input to the encoder. The autoencoder has one encoder and two decoders. 
One decoder (feature vector estimator) reconstructs the original input sample, while the other decoder (mask vector generator) learns to identify the inconsistency between feature values.


The second step is semi-supervised learning. The predictor uses the representation learned from pre-trained encoder. The predictor is trained in a semi-supervised manner using the consistency regularization loss. Several corrupted samples (augmented inputs) are used to compute the consistency loss for training the predictor.

\begin{figure}[t]
     \centering
         \includegraphics[width=0.49\textwidth]{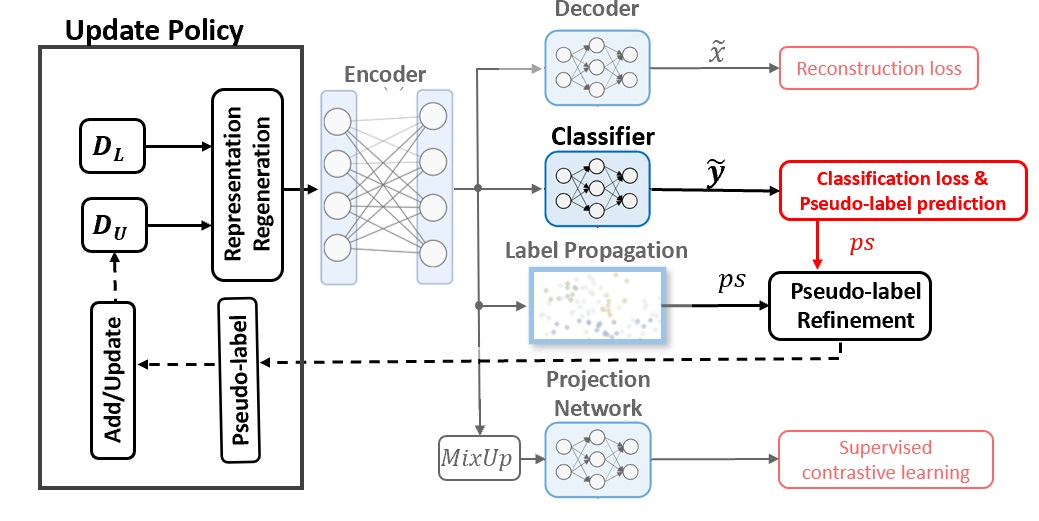}
         \label{fig:curls}
    \vspace*{-5mm}
        \caption{\emph{Progressive Contrastive Mixup} architecture. Three components are added to the architecture for training the encoder: classifier, pseudo-label refinement, and representation generation.}
    \vspace*{-3mm}
     \label{fig:ContrMixup}
\end{figure}

\subsubsection{Progressive VIME}~\\
We describe how the progressive architecture is added to \emph{VIME}. The black box in Figure \ref{fig:VIME}  shows  the \emph{Update Policy} Component. The \emph{Update Policy} is used to generate the CPR before feeding samples to the model.
 We introduce the term \emph{run} that stands for training the encoder and the predictor in both steps. So the first \emph{run} is equivalent to the original 
\emph{VIME}. 

In the beginning, the \emph{Update Policy} uses the labeled samples for generating representation because pseudo-labels for unlabeled samples are not available yet. The training follows two consecutive steps for the unsupervised training of autoencoder and semi-supervised training of the predictor, this is one \emph{run}. After training the predictor (blue box) in the second step of a \emph{run}, the pseudo-labels are generated from the predictor, which is then sent to \emph{Update Policy} component (dotted red arrow in Figure \ref{fig:VIME}).  The \emph{refinement} can be used inside the \emph{Update Policy}. It uses the pseudo-labels' confidence produced from the classifier for filtering. \emph{Refinement} will filter out unreliable pseudo-labels. Depending on whether using \emph{refinement} or not,  refined pseudo-labels or all pseudo-labels are added to the unlabeled samples set. The \emph{Representation Regeneration} in \emph{Update Policy} uses the labeled and unlabeled samples to generate the new representation for categorical features. When the representation is updated, the new training in the next \emph{run} is started on the latest representation of the data. In other word, the representation is updated before the next \emph{run}.

\subsubsection{Contrastive Mixup} ~\\
Before describing the progressive architecture, we review the \emph{Contrastive Mixup}\cite{darabi2021contrastive} first.
Contrastive Mixup leverages mixup-based augmentation in latent space for contrastive learning.
The \emph{Contrastive Mixup} has two training steps like \emph{VIME}. 
The first step is to train an autoencoder using labeled and unlabeled data by contrastive learning and reconstruction. 
The mixup operations are applied in the latent space on the samples with the same labels for contrastive learning. 
Note, unlike most mixup-based augmentation methods that they randomly select samples for mixup, \emph{Contrastive Mixup} select samples with the same label for mixup operation. 
Supervised contrastive learning is applied between the original and mixed samples. 
The label-propagation \cite{iscen2019label} is used after warm-up to generate or update pseudo-labels on unlabeled samples. 
Pseudo-labels will eventually be used for contrastive learning.
In the second step of the training, a predictor is trained using consistency loss as a regularization. The transparent components in Figure \ref{fig:ContrMixup} show the original architecture of \emph{Contrastive Mixup}. 
The black box component is proposed in progressive architecture.
 
\subsubsection{Progressive Contrastive Mixup}~\\
We introduce \emph{Progressive Contrastive Mixup} where \emph{Update Policy} and \emph{Refinement} are plugged into the \emph{Contrastive MixUp} \cite{darabi2021contrastive}. All components in Figure \ref{fig:ContrMixup} show the proposed progressive architecture. The highlighted components in Figure  \ref{fig:ContrMixup} show the changes compared to the original architecture.
We propose adding a classifier connected to the encoder to offer an alternative way to produce pseudo-labels as well as confidence on the pesudo-labels.

Like \emph{Progressive VIME}, we need to update the representation using \emph{Update Policy}. In \emph{Update Policy}, we can use all pseudo labels or a subset of them using \emph{Refinement} component. We believe that the \emph{Refinement} may improve the accuracy because of using more accurate pseudo-labels to generate the representation.

There are different strategies for performing \emph{Refinement} in \emph{Progressive Contrastive Mixup} because both label-propagation and the proposed classifier components are available in the architecture and can indicate the confidence on the pseudo-labels. We propose a two-step filtering mechanism (illustrated in \emph{Pseudo-label Refinement} component in Figure \ref{fig:ContrMixup}). Firstly, the pseudo-label agreement between the label-propagation and classifier is used to select more likely correct pseudo-labels. Then, the pseudo-labels are filtered based on a threshold on the weights measured by the label-propagation method. Based on our study, we find that label-propagation weights are more robust than the classifier's confidence scores.




%% file: experiment.tex
\section{Experimental Study}
This section shows the results of progressive architectures on different tabular datasets with the high-cardinality categorical data. Also, the progressive architecture is compared  with \emph{VIME} \cite{yoon2020vime} and \emph{Contrastive Mixup} \cite{darabi2021contrastive} on three datasets.  
A detailed study of the performance of using different components in  progressive architectures and the original architectures are presented.

\begin{table*}
\centering
\caption{Prediction accuracy of the progressive VIME on all Datasets ({Mean \textpm Standard deviations} are computed over 5 \emph{runs})}
\begin{tabular}{*{7}{lccc}}
  \toprule
  \textbf{Method} & \textbf{Drug Directory} & \textbf{Display Advertising} & \textbf{Traffic Violations} \\
  \midrule
  Supervised                       & 93.41\% (\textpm 0.3698)            & 70.44\% (\textpm 1.027)  & 77.932\% (\textpm 0.297) \\ 
  \hline
  VIME:Self-Supervised             & 91.61\% (\textpm 0.8044)            & 74.31\% (\textpm 0.458)  & 75.96\% (\textpm 2.794) \\
  VIME:Semi-Supervised             & 89.67\% (\textpm 0.2.507)           & 74.61\% (\textpm 0.414)  & 78.6\% (\textpm 0.47) \\
  \hline
  Progressive-VIME:Self-Supervised with update & 93.51\% (\textpm 0.5658)           & 73.166\% (\textpm 1.041)  & 78.89\% (\textpm 0.316) \\
  Progressive-VIME:Semi-Supervised with update & 92.86\% (\textpm 1.835)  & 74.788\% (\textpm 0.368) & 79.037\% (\textpm 0.283)\\
  Progressive-VIME: Self-Supervised with refinement & 93.638\%(\textpm 0.918) & 74.438\%(\textpm 0.861) & 78.77\% (\textpm 0.36)\\
  Progressive-VIME: Semi-Supervised with refinement & \textbf{94.72\%(\textpm 0.5128)} & \textbf{74.96\%(\textpm 0.226)} & \textbf{79.92\%(\textpm 0.504)}\\
  \bottomrule      
\end{tabular}
\label{tab:VIME-Results}
\vspace{-3mm}
\end{table*}

\begin{table*}
\centering
\caption{Prediction accuracy of the \emph{Contrastive Mixup} and \emph{Progressive Contrastive mixup} with \emph{Update Policy} and \emph{Refinement} on  all datasets ({Mean \textpm Standard deviations} are computed over 4 \emph{runs} with different seeds)}
\begin{tabular}{*{7}clcc}
\toprule

& \textbf{Method} & \textbf{Drug Directory} &  \textbf{Display Advertising} & \textbf{Traffic Violations} \\

\midrule
& Supervised & 93.33\%(\textpm 0.2872) &71.34\% (\textpm 0.541) & 77.64\% (\textpm 0.877) \\
\hline
& Contrastive Mixup & 93.47\%(\textpm 0.072) &71.25\%(\textpm 0.141) & 77.941\%(\textpm0.378) \\
\hline
\multirow{2}{*}{Without classifier} & Progressive Contrastive Mixup with update & 92.77\%(\textpm 0.7357) & 72.23\%(\textpm 1.12) & \textbf{78.75\% (\textpm 0.872)} \\
~ & Progressive Contrastive Mixup with refinement & 93.44\%(\textpm 0.5996)  & 70.85\% (\textpm 0.91) & 77.55\% (\textpm 0.694) \\
\hline
\multirow{2}{*}{With  Classifier} & Progressive Contrastive Mixup with update & 92.92\%(\textpm 0.8192)  &71.9\% (\textpm 0.71) & 76.93\% (\textpm 1.58) \\
~ & Progressive Contrastive Mixup with refinement & \textbf{94.44\%(\textpm 0.2384)}  & \textbf{75.01\% (\textpm 0.19)} & 78.29\% (\textpm 0.221) \\
\hline

\end{tabular}
\label{tab:contrastivemixup}
\vspace{-3mm}
\end{table*}

\subsection{Tabular Datasets}
To demonstrate the efficacy of the CPR, \emph{Update Policy} and \emph{Refinement} on the progressive architectures, we conduct a series of experiments on three datasets: Traffic Violations \cite{TrafficViolations}, Drug Directory \cite{DrugDirectory} and Display Advertising Challenge \cite{Criteo}.  All datasets have multiple high-cardinality categorical data. We shortly introduce each dataset and describe the cardinality of some features in the following. More details of the datasets are shown in the Table \ref{tab:datasets}.

The Traffic Violations dataset has 1,578,154 samples with multiple categorical features. The sum of the cardinality of all categorical features is higher than 200,000. This dataset contains DateTime, numerical, boolean, and string categorical features. We exclude DateTime features in our experiments.

The Display advertising challenge dataset is published by Criteo company. This dataset contains 11 numerical features (mostly count features) and 26 categorical features. The values of these features have been hashed onto 32 bits for anonymization purposes. One million samples are randomly sampled from the dataset for the experiment. This sampled dataset contains some categorical features that have cardinality higher than 200,000. 

The Drug Directory dataset is the smallest in our experiments. It has 19,764 samples with several categorical features. In this dataset, the feature with maximum cardinality has 5,032 unique values. This dataset contains the numerical, date, and categorical features. 

For preprocessing, the date features are converted to numerical values. All numerical features are normalized using standard scaler in scikit-learn library \footnote{\url{https://scikit-learn.org}}. Also, categorical features are converted to numerical features using the CPR.

\begin{table}
  \caption{The summary of datasets. The number of categorical and the number of non-categorical features in each dataset are shown. Most of the features are categorical.
More than one feature in each dataset has high cardinality}
  \label{tab:datasets}
  \footnotesize
  \begin{tabular}{p{0.23\linewidth} p{0.1\linewidth} p{0.12\linewidth} p{0.14\linewidth} p{0.14\linewidth} }
    \toprule
    \textbf{Dataset} & \textbf{\#cat cols} & \textbf{\#non-cat cols} & \textbf{\#Samples} & \textbf{Max cardinality}\\

    \midrule
    Drug Directory & 6 & 2 & 19,746 & 5,032\\
    Traffic Violations & 14 & 12 & 1,578,154 & 163,365\\
    Display Advertising Challenge & 26 & 11 & 1,000,000 & 321,439\\
  \bottomrule
\end{tabular}
\vspace{-5mm}
\end{table}

In the experiment, we use 80\% of the data as the train set and 20\% as the test set. Also, 10\% of the train set is chosen as the labeled set and the rest as the unlabeled set. This division of dataset is used in all experiments. The prediction accuracy on the test set is used as the metric for evaluation. We use the existing baselines and their codebases in \emph{VIME} and \emph{Contrastive Mixup} to perform all experiments. Also, we use the same network architecture and training protocol including the optimizer, learning rate schedule, etc.  The implementation of our proposed \emph{Progressive VIME} and \emph{Progressive Contrastive Mixup} can be found at GitHub page
\footnote{\url{https://github.com/mmoha014/Progressive_VIME_ContrastiveMixup}}.

\subsection{Experiments}
This section evaluates the proposed progressive architectures on all the aforementioned datasets. To explore the efficacy of our progressive semi-supervised framework on limited labeled data in practical setting, we compare the accuracy with the state-of-the-art methods by varying components in both proposed progressive architectures.

\subsubsection{Progressive VIME} ~\\
We report the performance of original \emph{VIME} method along with \emph{Progressive VIME} in Table \ref{tab:VIME-Results}. We evaluate the performance gain of each component in \emph{VIME} and \emph{Progressive VIME}. This evaluation shows how \emph{VIME} works on the CPR and how the \emph{Progressive VIME} performs versus \emph{VIME}. 

The components and the training methods that are shown in the Table \ref{tab:VIME-Results} for evaluation are described in the following:
\begin{itemize}
  \item \textbf{Supervised Model:} Train the predictor in the second step using the labeled data.
  
  \item \textbf{VIME:Self-Supervised:} The encoder is trained in step 1 and the predictor is trained in step 2. Only labeled data is used in the second step to train the predictor. 
  
  \item \textbf{VIME:Semi-Supervised:} Similar to self-supervised method, there are 2 steps in training. Difference is that in the second step, the data augmentation and consistency loss are used for training the predictor.

  \item \textbf{Progressive VIME:Self-supervised with update:} Adding \emph{Update Policy} and several \emph{run}s of training to the \emph{VIME:self-supervised} method. All pseudo-labels are used for updating the representation (CPR).
  
  \item \textbf{Progressive VIME:Semi-supervised with update:} Adding \emph{Update Policy} and several \emph{run}s of training to the \emph{VIME:semi-supervised} method. All pseudo-labels are used for updating the representation.
  
  \item \textbf{Progressive VIME:Self-supervised with refinement:} Adding \emph{Update Policy} and several \emph{run}s of training the \emph{VIME:self-supervised} method.  Updating the representation in \emph{Update Policy} component using more confident pseudo-labels in the predictor.
  \item \textbf{Progressive VIME:Semi-supervised with refinement:} Adding \emph{Update Policy} and several \emph{run}s of training to the \emph{VIME:semi-supervised} method. Selection of pseudo-labels based on the confidence of the predictor to update the representation in \emph{Update Policy} component. 
 \end{itemize}


\begin{table*}
\caption{Prediction accuracy of \emph{Contrastive Mixup} and \emph{Progressive Contrastive Mixup} containing different components for training the encoder on \textbf{Traffic Violation Dataset}. The cell with blue color show the best accuracy in the same row. The red color shows the best accuracy in the table.}
\footnotesize
{\begin{tabular}{|l|l|l|l|l|l|l|}
\hline
 $Method\downarrow \backslash Components \rightarrow$ & Classifier & Decoder & Classifier+Decoder & Classifier+Projection & Decoder+Projection & All components\\
\hline
 \textbf{Training without update}   & 77.72\% (\textpm 0.737) & \leavevmode\color{blue}\textbf{78.08\% (\textpm 0.034)} & 77.67\% (\textpm 0.467) & 77.72\% (\textpm 0.807) & 77.94\% (\textpm 0.378) & 77.94\% (\textpm 0.377)\\ 
\textbf{Training With Update}      & 77.05\% (\textpm 1.046) & 75.73\% (\textpm 1.53) & 76.6\% (\textpm 1.733) & 77.78\% (\textpm 0.707) & \leavevmode\color{green}\textbf{78.75\% (\textpm 0.872)} & 76.93\% (\textpm 1.58)  \\
\textbf{Training with Refinement}  & 78.063\% (\textpm 0.413) & 75.83\% (\textpm 0.745)  & 77.69\% (\textpm 1.01) & 77.93\% (\textpm 0.444) & 77.55\% (\textpm 0.694) & \leavevmode\color{blue}\textbf{78.29\% (\textpm 0.221)}\\
\hline
  \multicolumn{7}{|c|}{\textbf{Supervised}  77.64\% (\textpm 0.877)}\\
\cline{1-7}
  
\end{tabular}}
\label{tab:Mixup-Results_TVS}
\end{table*}




\begin{table*}
\caption{Prediction accuracy of \emph{Contrastive Mixup} and \emph{Progressive Contrastive Mixup} containing different components for training the encoder on \textbf{DisplayAdvertising Challenge Dataset}. The cell with blue color shows the best accuracy in the same row. The red color shows the best accuracy in the table.}
\footnotesize
{\begin{tabular}{|l|l|l|l|l|l|l|}
\hline
$Method\downarrow \backslash Components\rightarrow$   & Classifier & Decoder & Classifier+Decoder & Classifier+Projection & Decoder+Projection & All components\\
\hline
   \textbf{Training without update}   & 69.96\% (\textpm 2.32) & \leavevmode\color{blue}\textbf{71.884\% (\textpm 1.21)} & 70.135\% (\textpm 1.93) & 68.2\% (\textpm 2.2) & 71.25\% (\textpm 1.41) & 69.49\% (\textpm 0.534)\\ 
  \textbf{Training With Update}      & 71.365\% (\textpm 0.43) & 71.013\% (\textpm 0.252) & 71.52\% (\textpm 0.52) & 71.37\% (\textpm 0.373) & \leavevmode\color{blue}\textbf{72.23\% (\textpm 1.12)} & 71.9\% (\textpm 0.71)  \\

  \textbf{Training with Refinement}  & 74.91\% (\textpm 0.255) & 70.46\% (\textpm 1.72)  & 74.95\% (\textpm 0.262) & 74.68\% (\textpm 0.32) & 70.85\% (\textpm 0.91) & \leavevmode\color{green}\textbf{75.01\% (\textpm 0.19)}\\
 
  \hline
  \multicolumn{7}{|c|}{\textbf{Supervised} 71.34\% (\textpm 0.541)}\\
\cline{1-7}
\end{tabular}}
\label{tab:Mixup-Results_CDAS}
\end{table*}



\begin{table*}
\caption{Prediction accuracy of \emph{Contrastive Mixup} and \emph{Progressive Contrastive Mixup} containing different components for training the encoder on \textbf{Drug Directory Dataset}. The cell with blue color shows the best accuracy in the same row. The red color shows the best accuracy in the table.}
\footnotesize
{\begin{tabular}{|l|l|l|l|l|l|l|}
\hline
$Method\downarrow \backslash Components\rightarrow $  & Classifier & Decoder & Classifier+Decoder & Classifier+Projection & Decoder+Projection & All components\\
\hline
  \textbf{Training without update}   & 93.33\% (\textpm 0.6385) & 92.95\% (\textpm 0.403) & 93.4\% (\textpm 0.3534) & 93.18\% (\textpm 0.9351) & \leavevmode\color{blue}\textbf{93.47\% (\textpm 0.72)} & 93.31\% (\textpm 0.5687)\\ 
  
  \textbf{Training With Update}      & \leavevmode\color{blue}\textbf{93.22\% (\textpm 0.7485)} & 92.32\% (\textpm 0.68) & 92.74\% (\textpm 0.7999) & 92.715\% (\textpm 0.9141) & 92.77\% (\textpm 0.7357) & 92.92\% (\textpm 0.8192)  \\
  
  \textbf{Training with Refinement}  & 93.98\% (\textpm 0.5288) & 93.55\% (\textpm 0.615)  & \leavevmode\color{green}\textbf{94.28\% (\textpm 0.2022)} & 94.13\% (\textpm 0.6028) & 93.44\% (\textpm 0.5996) & 94.24\% (\textpm 0.2384)\\
  \hline
  \multicolumn{7}{|c|}{\textbf{Supervised}  93.33\% (\textpm 0.2872)}\\
\cline{1-7}
\end{tabular}}\label{tab:Mixup-Results_DDS}
\end{table*}

  

Table \ref{tab:VIME-Results} shows the proposed \emph{Progressive VIME with refinement} outperforms the other methods, resulting in the best prediction performance. The \emph{VIME} underperforms on \emph{Drug directory} dataset in comparison with supervised method, while progressive method is more robust and provide consistent improvement on all datasets. In other words,  the progressive methods with update and refinement always achieve better performance compared with the original \emph{VIME} and supervised methods. The higher model's predictive power shows the advantage of the progressive approach in leveraging the unlabeled data and learning better representations.  The \emph{progressive training} with refinement performs slightly better than \emph{progressive training} with update because it only keeps likely correct pseudo-labels, which results in better representation regeneration.


\subsubsection{Progressive Contrastive Mixup} ~\\
In this section, we compare our progressive architecture with the \emph{Contrastive Mixup} using the same aforementioned datasets. We compare the performance of different models. All the models used for comparison are introduced as follows.
\begin{itemize}
    \item \textbf{Supervised Model:} using predictor in the second step and training it on the labeled samples.
    \item \textbf{Contrastive Mixup:} training the original \emph{Contrasive Mixup} without updating the CPR. Training the encoder in the first step, then training predictor in the second step.
    \item \textbf{Progressive Contrastive Mixup with update:} This method adds \emph{Update Policy} to \emph{Contrastive Mixup}  and updates the representation using all pseudo-labels. When the classifier is used, it just effects in training of the representation and do not participate in  updating the CPR.
    \item \textbf{Progressive Contrastive Mixup with refinement:} This method adds \emph{Update Policy} and the pseudo-label refinement components to the \emph{Contrastive Mixup}. 
    \begin{itemize}
    \item For \textbf{without classifier} method, the pseudo-label refinement component filters the pseudo-labels only based on a threshold on the weights calculated by the label-propagation method. 
    \item For \textbf{with classifier} method, refining the pseudo-labels on the weights of the label-propagation method is used after the pseudo-labels agreement. In other words, the pseudo-labels are refined twice. 
    \end{itemize}
\end{itemize}

Table \ref{tab:contrastivemixup} shows that the progressive training outperforms the original \emph{Contrastive Mixup} \cite{darabi2021contrastive} and supervised models. 
On \emph{Display Advertising Challenge} and \emph{Drug Directory} datasets, \emph{Refinement} using classifier in the architecture performs the best, while on \emph{Traffic Violations} dataset the best performance obtained by the proposed progressive training without a classifier and \emph{Refinement}. 

The results on the \emph{Display Advertising Challenge} dataset show that \emph{Contrastive Mixup} does not always perform better than the supervised method, but progressive training outperforms the supervised method.

Finally, both Tables \ref{tab:VIME-Results} and \ref{tab:contrastivemixup} show that \emph{VIME} and \emph{Contrastive Mixup} can not consistently improve the accuracy compared to the supervised approach on all datasets, while the progressive training outperforms the supervised,  and non-progressive methods on all datasets. 

\subsubsection{Ablation Study} ~\\
We evaluate the \emph{Update Policy}, \emph{Refinement} and other components in \emph{Progressive Contrastive Mixup}. The effect of each component (Classifier, Projection Network, Decode, and Label-propagation) is studied to examine how they affect final performance. Components changes are only made in the first training step, and the second step stays the same. 
Because the label-propagation is an important part of the \emph{Contrastive Mixup}, it is used in all evaluations. 

We use two terms: \emph{with update} and \emph{with refinement}. The \emph{with update} is used when we add the \emph{update policy} based on all pseudo-labels without refinement. The term \emph{with refinement} means we use both \emph{update policy} and \emph{refinement}. In contrast, when we do not use these terms \emph{with update} and \emph{with refinement}, it means that they are not used and there is no regenerating data representation in the architecture. 
Two-step refinement mechanism is used in this experiment, where threshold on the weights of pseudo-labels generated by the label-propagation is set to 0.9 in our experiment.

We want to study the robustness of the propose framework under various combinations of the components.
Mainly three different cases are compared: 1. without updating CPR; 2. updating CPR with all pseudo-labels; 3. refining pseudo-labels for updating.
Table \ref{tab:Mixup-Results_TVS},  \ref{tab:Mixup-Results_CDAS} and \ref{tab:Mixup-Results_DDS} show the results of this study in Traffic Violations, Display Advertising Challenge and Drug Directory datasets. 
Note that, \textbf{Projection+decoder} architecture is the architecture proposed in \emph{Contrastive Mixup}.

Three Tables ~\ref{tab:Mixup-Results_TVS}, ~\ref{tab:Mixup-Results_CDAS} and ~\ref{tab:Mixup-Results_DDS} show that updating the conditional probability representation consistently outperforms the case when conditional probability representation is not updated across different models. 
It demonstrate the robustness of the proposed framework.
In all three datasets, when refinement is added, \textbf{Training with Refinement} outperforms the \textbf{Training without update} and \textbf{Training with update} in majority cases. 
This shows that \textbf{Training with Refinement} uses more accurate pseudo-labels and improves the data representation regeneration.
In \emph{Traffic Violations} dataset, adding \textbf{Training with update} to the \emph{Contrastive Mixup} performs best. 
In \emph{Display Advertising Challenge} dataset, adding the proposed \emph{Classifier} and \textbf{Training with Refinement} achieves the best result, which improves \emph{Contrastive Mixup} by $5.28\%$.
The proposed framework achieves the highest improvement in \emph{Display Advertising Challenge} dataset because cardinality of categorical features are really high.
In \emph{Drug Directory} dataset, the best performing method adds the proposed \emph{Classifier} and \textbf{Training with Refinement} but removes the contrastive learning component.
In both \emph{Display Advertising Challenge} and \emph{Drug Directory} datasets, the proposed classifier brings positive effective to the model architecture.
In \emph{Traffic Violation} and \emph{Drug Directory} dataset, we observe that \textbf{Training with update} underperforms \textbf{Training without update} probability because the noise introduced by wrong pseudo-labels.

%% file: conclusion.tex
\section{Conclusion}
In this paper, we advocate rethinking the semi-supervised learning problem on tabular domain from the feature representation perspective, especially for high-cardinality categorical data.
dInstead of sticking to the most popular representation (one-hot encoding, label encoding), we propose using conditional probability representation and keep upgrading the representation during training.
Upgrading representation is realized by leveraging labels for labeled data and pseudo-labels for unlabeled data.
Refinement mechanism is also proposed to reduce the noise introduced to the feature representation through pseudo-labels.
We demonstrate the effectiveness and robustness of the proposed framework by incorporating it with the different algorithms and evaluating it on different datasets.
Note that, the progressive updating framework proposed in this work is not contradictory to the existing semi-supervised learning approaches, but complementary to help gain more understanding of the problem.
We hope with the awareness of many encoding tools for tabular domain data, it becomes easier to learn meaningful representations in tabular domain.

%% file: supplementary/supp.tex
\section{Appendix}
\subsection{Experimentation setup details}

The \emph{Progressive Contrstive Mixup} uses refinement threshold 0.9 on the weights of the label-propagation algorithm and the \emph{Progressive VIME} uses different threshold in range [0.7, 0.9] for confidence of classifier in each dataset. In \emph{Progressive Contrastive Mixup}, the weight of the loss for the classifier is 0.5.


All the setup follows what is used in the original papers (including all hyperparameters for loss functions, perturbation probability in mask generator of \emph{VIME}, number of nearest neighbors in the label-propagation method of \emph{VIME} and \emph{Contrastive Mixup}).
Except, for \emph{VIME} related architectures, the size of the latent representation in the encoder is changed. We assigned number of dimensions as 46, 42, and 64 on Traffic Violations, Drug Directory, and Display Advertising Challenge datasets respectively. For reproducible results in \emph{Contrastive Mixup}, we use random seed number 123,127,131,137.

All components except the label-propagation and \emph{Update Policy} use fully connect layers like the settings in \emph{VIME} and \emph{Contrastive Mixup}.


\section{Additional experiments}
\begin{figure}[h!]
     \centering
         \includegraphics[width=0.45\textwidth]{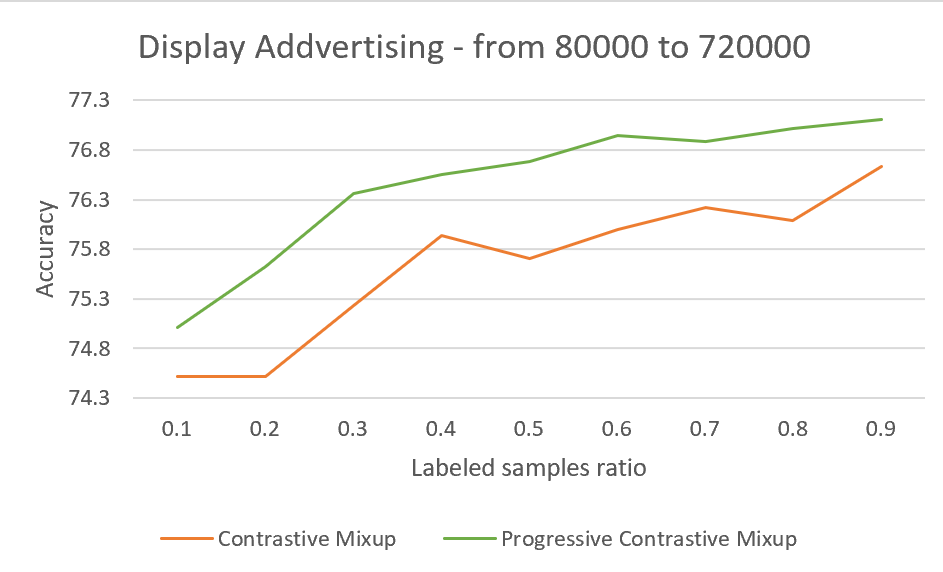}
         \label{fig:curls}
    \vspace*{-3mm}
        \caption{Comparison of performance on Display Advertising Challenge Dataset under varying labeled sample ratios on \emph{Progressive Contrastive Mixup} and \emph{Contrastive Mixup}. The range of the x-axis is given as [0.1, .9]. }
     \label{fig:ratio_criteo_mixup}
     \vspace{-5mm}
\end{figure}

\begin{figure}[h!]
     \centering
         \includegraphics[width=0.45\textwidth]{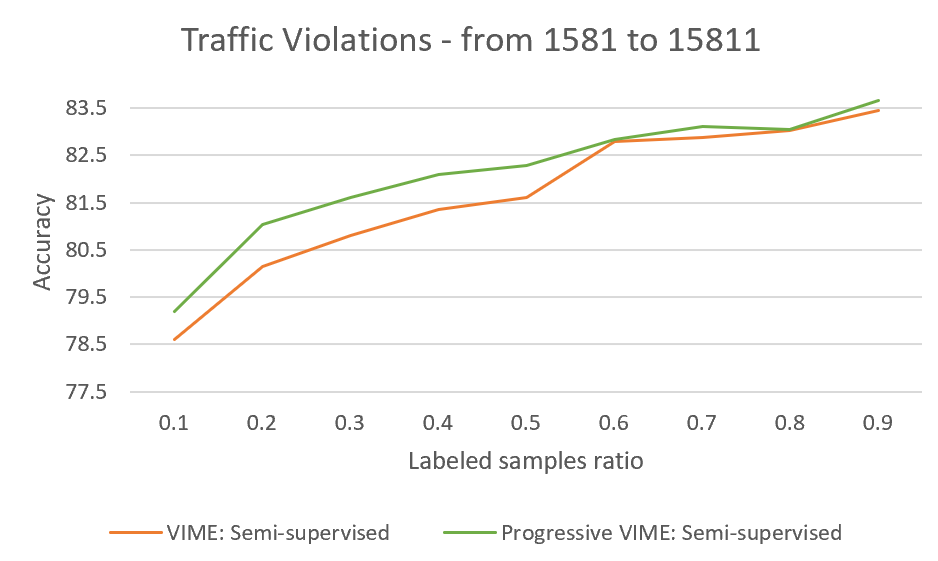}
         \label{fig:curls}
    \vspace*{-3mm}
        \caption{Comparison  of the performance on Traffic Violations Dataset under varying labeled sample ratios on \emph{Progressive VIME} and \emph{VIME}. The range of the x-axis is given as [0.1, .9]}
     \label{fig:ratio_traffvio_vime}
     \vspace{-3mm}
\end{figure}

\subsection{Additional Experiments}
\subsubsection{Performance with different ratios of labeled samples}
We have performed the additional experiments to verify the effectiveness of the progressive method under varying number of labeled samples. The \emph{Progressive Contrastive Mixup} on the Display Advertising Challenge dataset and \emph{Progressive Mixup} on the Traffic Violations dataset are used to show the results. 

Figure \ref{fig:ratio_criteo_mixup} shows that under different ratios for labeled samples the \emph{Progressive Contrastive Mixup} outperforms \emph{Contrastive Mixup} on Display Advertising Challenge dataset. Figure \ref{fig:ratio_traffvio_vime} demonstrates the effectiveness of \emph{Progressive VIME} on Traffic Violations dataset on the majority of different ratios of the labeled samples. When less labeled sample are used for training (ratio smaller than 0.5), \emph{Progress VIME} clearly outperforms \emph{VIME}.
When more labeled samples are involved in the training (ratio larger than 0.5), \emph{VIME} and \emph{Progress VIME} perform closely. Possible reason could be that 50\% of data can already provide accurate estimation of real CPR, any additional data can not bring additional value for finding more accurate representations. 
It might be the reason for close performance on the Traffic Violations dataset. 
Overall, the proposed progressive method in most experiments on both datasets outperforms the non-progressive methods.





\begin{table}
\centering
\caption{Prediction accuracy of the \emph{Contrastive Mixup} and \emph{Progressive Contrastive mixup} with \emph{Update Policy} and \emph{Refinement} on  Display Advertising Challenge and Traffic Violations datasets }

\begin{tabular}{>\centering m{3cm}cc}
\toprule
 \textbf{Method}  &  \textbf{Display Advertising} & \textbf{Traffic Violations} \\

\midrule
 Supervised & 75.66\%(\textpm 0.11) & 79.84\% (\textpm 0.13) \\
\hline
Contrastive Mixup & 75.66\% (\textpm 0.12) & 80.45\% (\textpm 0.232) \\
\hline
Progressive Contrastive Mixup with update (no classifier) & 77.18\% (\textpm 0.078) & \textbf{81.53\% (\textpm 0.103)} \\
Progressive Contrastive Mixup with refinement (no classifier) & 77.13 (\textpm 0.065) & 80.32\% (\textpm 0.122) \\
\hline
Progressive Contrastive Mixup with update and classifier & \textbf{77.21\% (\textpm 0.058)} & 80.65 (\textpm 0.33) \\
Progressive Contrastive Mixup with refinement and classifier & 77.08\% (\textpm 0.01) & 80.43\% (\textpm 0.212) \\
\hline
\end{tabular}
\label{tab:AppTE}
\vspace{-3mm}
\end{table}

\subsubsection{Performance on Target Encoding}
The proposed framework is not restricted to only conditional probability representation. It is flexible as long as the representation method has baked target label information into the efficient representation. One other encoding method called target encoding (Target Encoding \cite{TE2001}) is very similar to CPR since it also considers target label in the representation. Additional experiments in Table \ref{tab:AppTE} have been performed to study the performance of Target Encoding \cite{TE2001}) in semi-supervised learning problems, similar effect has been observed.